\documentclass[11pt]{article}
\usepackage{amsmath}
\usepackage{acl}
\usepackage{times}
\usepackage{latexsym}
\usepackage{graphicx}
\usepackage{booktabs}
\usepackage{hyperref}
\usepackage{url}
\usepackage{microtype}
\usepackage{forest}
\usepackage{forest}
\usepackage{tikz}
\usetikzlibrary{arrows.meta, shapes, positioning}

\tikzset{
  circle arrow/.style={
    draw,
    -{Circle[open,length=2pt]},
    line width=0.5pt
  }
}
\usepackage{forest}
\usepackage{xcolor}
\usepackage{hyperref}
\usepackage{graphicx}
\usepackage{float}
\usepackage{pgfplots}
\usepackage{caption}
\pgfplotsset{compat=1.18}
\usepackage{microtype}

\hypersetup{
    colorlinks=true,
    linkcolor=blue,
    urlcolor=blue,
    citecolor=blue
}
\forestset{
  intermediate/.style={font=\tiny\sffamily},
}

\newcommand{\corrauthor}{\thanks{Corresponding author}}

\title{Federated Retrieval-Augmented Generation: A Systematic Mapping Study}

\author{%
  \textbf{Abhijit Chakraborty}\corrauthor\textsuperscript{1} \and
  \textbf{Chahana Dahal}\textsuperscript{2} \and
  \textbf{Vivek Gupta}\footnotemark[1]\textsuperscript{1} \\[4pt]
  \textsuperscript{1}Arizona State University \qquad
  \textsuperscript{2}University of Nevada, Las Vegas \\[2pt]
  {\tt \{achakr40,vgupt140\}@asu.edu} \tt \{chahana.dahal\}@unlv.edu
}

\date{}

\begin{document}
\maketitle

\begin{abstract}
Federated Retrieval-Augmented Generation (Federated RAG) combines Federated Learning (FL),which enables distributed model training without exposing raw data, with Retrieval-Augmented Generation (RAG), which improves the factual accuracy of language models by grounding outputs in external knowledge. As large language models are increasingly deployed in privacy-sensitive domains such as healthcare, finance, and personalized assistance, Federated RAG offers a promising framework for secure, knowledge-intensive natural language processing (NLP). To the best of our knowledge, this paper presents the first systematic mapping study of Federated RAG, covering literature published between 2020 and 2025. Following Kitchenham’s guidelines for evidence-based software engineering, we develop a structured classification of research focuses, contribution types, and application domains. We analyze architectural patterns, temporal trends, and key challenges, including privacy-preserving retrieval, cross-client heterogeneity, and evaluation limitations. Our findings synthesize a rapidly evolving body of research, identify recurring design patterns, and surface open questions, providing a foundation for future work at the intersection of RAG and federated systems.
\end{abstract}
\section{Introduction}
Large language models (LLMs) are increasingly deployed in domains such as healthcare, finance, and personalized assistance, raising concerns about data privacy, ownership, and regulatory compliance. In response, \textbf{Federated Learning} (FL) has emerged as a compelling paradigm for training models across distributed clients without exchanging raw data \citep{kairouz2021}. In parallel, \textbf{Retrieval-Augmented Generation} (RAG) \citep{lewis2020rag} enhances LLMs by grounding their outputs in dynamically retrieved external knowledge, reducing hallucinations and improving the accuracy of facts \citep{lewis2020}.


FL trains models across clients without sharing raw data, using methods like FedAvg~\citep{mcmahan2017} and secure aggregation, widely applied in privacy-sensitive~\citep{kairouz2021} domains.

RAG is a hybrid method that combines a text retriever with a generator. Before producing a response, the model first retrieves relevant text from an external knowledge base or document store. This process helps reduce hallucinations and improves factual consistency in the generated output. Popular implementations include the original RAG model \citep{lewis2020} and Fusion-in-Decoder (FiD) \citep{izacard2020}, which demonstrate how integrating retrieval can enhance open-domain question answering.

At the intersection of these paradigms lies \textbf{Federated RAG}, a hybrid approach that enables LLMs to access distributed knowledge sources in a privacy-preserving manner. It combines the strengths of data localization, personalized retrieval, and context-aware generation. Conceptually, Federated RAG builds on earlier federated search methods \citep{shokouhi2011}, which aggregated results from siloed sources without centralized indexing, and extends them to support complex generative tasks.

\textbf{Federated RAG} is distinctive because federated learning safeguards training-time privacy by restricting data to local silos, while retrieval-augmented generation grounds inference-time outputs in external evidence, reducing hallucinations. Their integration enables capabilities such as source attribution, local index maintenance, and dynamic document unlearning that neither paradigm achieves alone.

As LLMs have grown in capability, driven by innovations such as the Transformer architecture \citep{vaswani2017attention}, pre-training methods \citep{devlin2019bert,radford2019gpt2,brown2020gpt3}, and prompting strategies \citep{schick2021fewshot,liu2021pretrainprompt,debnath2025comprehensive}, so has the demand for integrating them with heterogeneous private data contexts. Since 2019, the parallel evolution of LLMs, RAG, and FL has laid the groundwork for Federated RAG to emerge as both a viable and increasingly necessary framework.

In this paper, we present the first systematic mapping study of Federated RAG, following the methodology of evidence-based software engineering proposed by Kitchenham \citep{kitchenham2011}. Our objective is to map how retrieval-augmented generation is being adapted and deployed across federated architectures, and to surface recurring trends, gaps, and design patterns in the literature. To this end, we address the following research questions:

\begin{itemize}
    \item \textbf{RQ1:} What are the dominant architectural patterns used to integrate Retrieval-Augmented Generation (RAG) into federated systems?
    \item \textbf{RQ2:} What are the primary research focuses and contribution types in the literature on federated RAG systems?
    \item \textbf{RQ3:} In which application domains (e.g., healthcare, finance, education) has federated RAG been explored, and what problems are these systems designed to solve?
    \item \textbf{RQ4:} What key challenges, open issues, and underexplored areas emerge from current research at the intersection of RAG and federated systems?
\end{itemize}

These questions structure our analysis of the literature published between 2020 and 2025, allowing us to classify prior work, identify underexplored areas, and provide a foundation for future research at the intersection of privacy, retrieval, and generation.
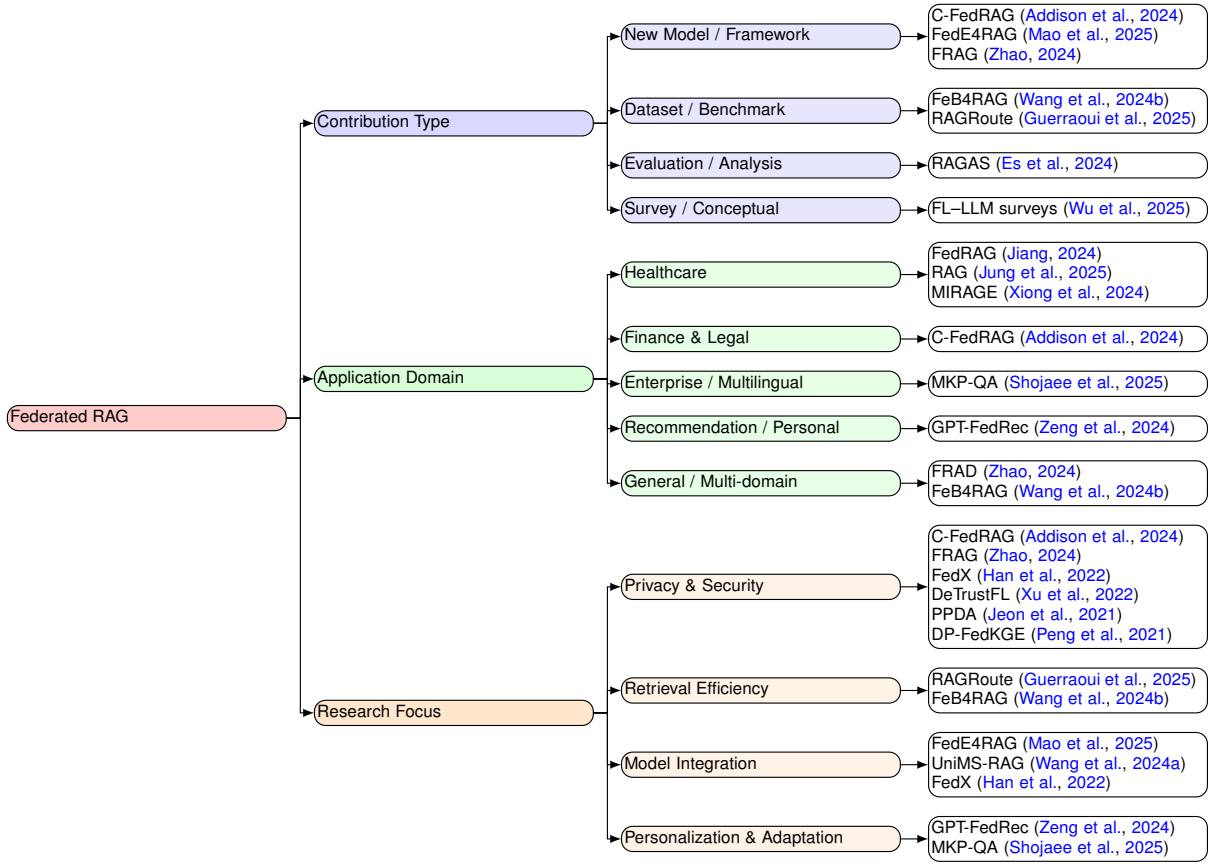
\begin{figure*}[t]
\centering
\resizebox{\textwidth}{!}{
\begin{forest}
for tree={
  grow'=0,
  draw,
  rounded corners,
  text width=3.4cm,
  align=left,
  l sep=10pt,
  s sep=7pt,
  inner sep=1pt,
  outer sep=0pt,
  minimum height=0pt,
  minimum width=0pt,
  edge path={
    \noexpand\path [draw, ->, >={latex}] (!u.parent anchor) -- +(5pt,0) |- (.child anchor)\forestoption{edge label};
  },
  parent anchor=east,
  child anchor=west,
  anchor=west,
  font=\scriptsize\sffamily\tiny,
  tier/.option={tier},
}
[Federated RAG, fill=red!20
  [Contribution Type, fill=blue!15, tier=intermediate
    [New Model / Framework, fill=blue!10
      [{\scriptsize \sffamily\tiny C-FedRAG~\citep{addison2024c} \\ FedE4RAG~\citep{shin2023} \\ FRAG~\citep{zhao2024}}, align=left]
    ]
    [Dataset / Benchmark, fill=blue!10
      [{\scriptsize \sffamily\tiny FeB4RAG~\citep{wang2024} \\ RAGRoute~\citep{guerraoui2025efficient}}, align=left]
    ]
    [Evaluation / Analysis, fill=blue!10
      [{\scriptsize \sffamily\tiny RAGAS~\citep{shah2023}}, align=left]
    ]
    [Survey / Conceptual, fill=blue!10
      [{\scriptsize \sffamily\tiny FL–LLM surveys~\citep{wu2025survey}}, align=left]
    ]
  ]
  [Application Domain, fill=green!15, tier=intermediate
    [Healthcare, fill=green!10
      [{\scriptsize \sffamily\tiny FedRAG~\citep{jiang2024clinical} \\ RAG~\citep{jung2025federatedlearningragintegration} \\ MIRAGE~\citep{xiong-etal-2024-benchmarking}}, align=left]
    ]
    [Finance \& Legal, fill=green!10
      [{\scriptsize \sffamily\tiny C-FedRAG~\citep{addison2024c}}, align=left]
    ]
    [Enterprise / Multilingual, fill=green!10
      [{\scriptsize \sffamily\tiny MKP-QA~\citep{shojaee2025federatedretrievalaugmentedgeneration}}, align=left]
    ]
    [Recommendation / Personal, fill=green!10
      [{\scriptsize \sffamily\tiny GPT-FedRec~\citep{zeng2024federatedrecommendationhybridretrieval}}, align=left]
    ]
    [General / Multi-domain, fill=green!10
      [{\scriptsize \sffamily\tiny FRAD~\citep{zhao2024} \\ FeB4RAG~\citep{wang2024}}, align=left]
    ]
  ]
  [Research Focus, fill=orange!20, tier=intermediate
    [Privacy \& Security, fill=orange!10
      [{\scriptsize \sffamily\tiny C-FedRAG~\citep{addison2024c} \\ FRAG~\citep{zhao2024} \\ FedX~\citep{han2022fedx} \\ DeTrustFL~\citep{xu2022detrust} \\ PPDA~\citep{9484437} \\ DP-FedKGE~\citep{peng2021differentially}}, align=left]
    ]
    [Retrieval Efficiency, fill=orange!10
      [{\scriptsize \sffamily\tiny RAGRoute~\citep{guerraoui2025efficient} \\ FeB4RAG~\citep{wang2024}}, align=left]
    ]
    [Model Integration, fill=orange!10
      [{\scriptsize \sffamily\tiny FedE4RAG~\citep{shin2023} \\ UniMS-RAG~\citep{chen2022} \\ FedX~\citep{han2022fedx}}, align=left]
    ]
    [Personalization \& Adaptation, fill=orange!10
      [{\scriptsize \sffamily\tiny GPT-FedRec~\citep{zeng2024federatedrecommendationhybridretrieval} \\ MKP-QA~\citep{shojaee2025federatedretrievalaugmentedgeneration}}, align=left]
    ]
  ]
]
\end{forest}
}
\caption{\textbf{Classification Scheme:} Summarizes the scheme with example categories and the distribution of studies (one study may fall into multiple categories).}
\label{fig:unified-classification}
\end{figure*}
\section{Methodology}
We searched the best NLP, ML and security venues (2020-2025) using terms such as \textit{ ``federated learning''}, \textit{``retrieval-augmented generation''} and \textit{ ``federated search''}, including backward references. From ~50 papers, 18 met criteria. Following Kitchenham’s~\citep{kitchenham2011} method, we coded each by (a) research focus, (b) contribution type, (c) application domain.

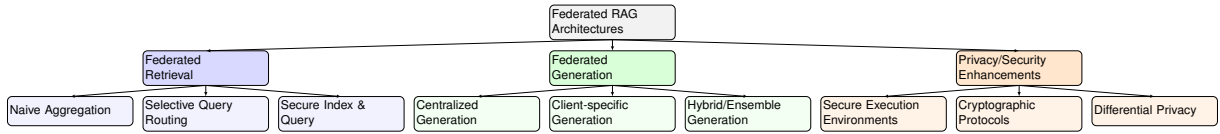
\begin{figure*}[t]
\centering
\resizebox{\textwidth}{!}{
\begin{forest}
for tree={
  grow=-90,
  draw,
  rounded corners,
  minimum height=2.4em,
  text width=3.8cm,
  font=\sffamily,
  anchor=north,
  parent anchor=south,
  child anchor=north,
  align=left,
  l sep=10pt,
  s sep=9pt,
  inner sep=2pt,
  outer sep=0pt,
  edge path={
    \noexpand\path [draw, ->, >={latex}]
    (!u.parent anchor) -- (.child anchor)\forestoption{edge label};
  },
}
[Federated RAG\\Architectures, draw, fill=gray!10
  [Federated\\Retrieval, draw, fill=blue!15
    [Naive Aggregation, draw, fill=blue!5]
    [Selective Query\\Routing, draw, fill=blue!5]
    [Secure Index \&\\Query, draw, fill=blue!5]
  ]
  [Federated\\Generation, draw, fill=green!15
    [Centralized\\Generation, draw, fill=green!5]
    [Client-specific\\Generation, draw, fill=green!5]
    [Hybrid/Ensemble\\Generation, draw, fill=green!5]
  ]
  [Privacy/Security\\Enhancements, draw, fill=orange!20
    [Secure Execution\\Environments, draw, fill=orange!10]
    [Cryptographic\\Protocols, draw, fill=orange!10]
    [Differential Privacy, draw, fill=orange!10]
  ]
]
\end{forest}
}
\caption{\textbf{Taxonomy} of Federated RAG Architectures. A conceptual map to classify the diverse system designs in the literature.}
\label{fig:federatedrag-architecture-taxonomy}
\end{figure*}

We also recorded the publication year of each study to observe temporal trends and emerging research fronts. Based on this categorization, we developed the unified classification scheme (Figure~\ref{fig:unified-classification}) to summarize the distribution of studies across focus areas and contribution types. Furthermore, this schema informed the design of our architectural taxonomy (Figure~\ref{fig:federatedrag-architecture-taxonomy}), which visually organizes the Federated RAG landscape by contribution, application, and system goals.
\section{Analysis of Recent Studies}
Retrieval ranges from naive aggregation to secure encrypted search (e.g., FRAG~\citep{zhao2024}). Generation spans centralized, client-specific, or hybrid (e.g., GPT-FedRec~\citep{zeng2024federatedrecommendationhybridretrieval}). These choices balance scalability, personalization, and privacy (Figure \ref{fig:federatedrag-architecture-taxonomy}).


These architectural choices require balancing scalability, performance, personalization, and privacy.
\subsection{Research Focus}
Foundational privacy work from 2020–2022 paved the way for Federated RAG.  
\textit{PPDA}~\citep{9484437} introduced lightweight decentralized aggregation without differential privacy (DP) or homomorphic encryption (HE).  
\textit{DP\mbox{-}FedKGE}~\citep{peng2021differentially} added entity-level DP for federated knowledge-graph embedding.  
\textit{DeTrustFL}~\citep{xu2022detrust} used cryptographic consensus to block disaggregation attacks, and  
\textit{FedX}~\citep{han2022fedx} employed cross-client distillation for unsupervised representation learning.

From 2023 through 2025, privacy mechanisms became system-level.  
\textit{C\mbox{-}FedRAG}~\citep{addison2024c} executes retrieval \emph{and} generation entirely inside trusted enclaves, securing clinical QA on \textit{MKP-QA}.  
\textit{FRAG}~\citep{zhao2024} realises IND-CPA-secure $k$NN search via single-key HE while keeping latency practical.

Parallel work targets efficiency. \textit{RAGRoute}~\citep{guerraoui2025efficient} learns to route queries to high-utility silos, cutting 75 \% redundant traffic on \textsc{MIRAGE}.  
\textit{FedB4RAG}~\citep{wang2024} benchmarks routing policies over 16 \textsc{BEIR} datasets, exposing accuracy–cost trade-offs.

Integration efforts replace isolated modules with end-to-end federated stacks.  
\textit{FedE4RAG}~\citep{shin2023} trains dense retrievers collaboratively, combining HE-protected parameter exchange with distillation to align heterogeneous embeddings.  
\textit{UniMS\mbox{-}RAG} unifies client-side retrieval with a shared generator, allowing adaptive control over privacy and latency.
\begin{table}[t]
\centering\small
\begin{tabular}{l c}
\toprule
\textbf{Research Focus} & \textbf{\# of Studies} \\
\midrule
Privacy \& Security & 6 \\
Personalization \& Adaptation & 2 \\
Retrieval Efficiency & 2 \\
Model Integration & 3 \\
\hline
\end{tabular}
\caption{\small Classification of primary research focus in federated RAG studies, aligned with the taxonomy shown in Figure~\ref{fig:unified-classification}.}
\label{tab:focus}
\vspace{-1.9em}
\end{table}
Table~\ref{tab:focus} summarizes these trajectories in the primary studies \emph{18} on map.  
Half still prioritize \emph{privacy and security}, reflecting the FL heritage of the field.  
\emph{Personalisation} (e.g., \textit{GPT\mbox{-}FedRec}, \textit{MKP\mbox{-}QA}) and \emph{retrieval efficiency} (e.g., \textit{RAGRoute}, \textit{FedB4RAG}) are rising, while \emph{model-level integration} (\textit{FedE4RAG}, \textit{UniMS\mbox{-}RAG}) signals a shift toward deployable cross-silo systems.

Overall, research has progressed from algorithmic proofs of concept (2020–2022) to holistic architectures (2023–2025) that simultaneously safeguard data, route queries intelligently, and harmonise retrieval with generation, marking a decisive step toward scalable Federated RAG. To complement these qualitative descriptions, Table \ref{tab:fedrag-allstudies} in the Appendix synthesizes empirical metrics reported across representative systems, highlighting trade-offs between privacy, efficiency, and personalization.

\subsection{Contribution Type}\label{para:contribution}\
The literatures on Federated RAG reflect a balanced evolution across four primary contribution types: \textit{new models or frameworks}, \textit{datasets and benchmarks}, \textit{evaluation and analysis tools}, and \textit{conceptual overviews}. As shown in Table~\ref{tab:contrib}, a substantial portion of the work focuses on developing novel system architectures for privacy-preserving generation and retrieval. \textit{FedE4RAG}\citep{shin2023} introduces a modular design that integrates homomorphic encryption and knowledge distillation to train federated dense retrievers, offering a decentralized alternative to traditional centralized systems. \textit{C-FedRAG}\citep{addison2024c} builds a confidential execution pipeline using Trusted Execution Environments (TEEs), enabling secure generation and retrieval particularly suited to sensitive domains like healthcare. \textit{FRAG}~\citep{zhao2024} contributes a cryptographic kNN retrieval system backed by IND-CPA-secure homomorphic encryption, facilitating privacy-preserving vector search.

Beyond system-level proposals, several contributions focus on datasets and benchmarking. \textit{FeB4RAG}\citep{wang2024} presents the first federated retrieval benchmark across 16 BEIR-derived domains, enabling rigorous evaluation of routing strategies and retrieval quality under resource constraints. \textit{MKP-QA}\citep{shojaee2025federatedretrievalaugmentedgeneration} introduces a domain-specific benchmark for multilingual enterprise QA, designed to assess the selection of cross-silo documents and the preservation of context. Additionally, \textit{RAGRoute}~\citep{guerraoui2025efficient} offers a specialized evaluation testbed for its adaptive query planning mechanism, allowing performance measurement across silos under constrained budgets.
\begin{table}[t]
\centering\small
\begin{tabular}{l r}
\toprule
\textbf{Contribution Type} & \textbf{\# of Studies} \\
\midrule
New Model / Framework     & 3 \\
Dataset / Benchmark       & 2 \\
Evaluation / Analysis     & 1 \\
Survey / Conceptual       & 1 \\
\bottomrule
\end{tabular}
\caption{\small Categorization of studies by primary contribution type. Majority propose new models; others contribute benchmarks, evaluation metrics, or conceptual overviews.}
\label{tab:contrib}
\vspace{-1.9em}
\end{table}

Evaluation and analysis contributions aim to establish consistent standards for measuring retrieval and generation quality in federated setups. \textit{RAGAS}~\citep{shah2023} proposes a suite of evaluation metrics tailored to federated RAG pipelines, accounting for inconsistencies in retrieved evidence, hallucinations, and robustness to noise injection. \textit{FeB4RAG} extends this line of work by benchmarking retrieval degradation when shifting from naive to intelligent routing policies, illustrating the tradeoffs between accuracy and resource consumption.

Finally, \textit{survey and conceptual} literature remains sparse. The only entry in this category is \textit{FL-LLM Surveys}~\citep{wu2025survey}, which provides a broader discussion on federated LLMs and briefly references RAG mechanisms. In contrast, our mapping study represents to the best of our knowledge, the first structured synthesis of architectural, empirical, and deployment-oriented work in Federated RAG, addressing a clear gap in the literature.

\subsection{Application Domains and Use Cases}
Federated RAG is gaining traction across a variety of domains that require both strong privacy guarantees and access to distributed knowledge sources. Among these, \textit{healthcare} has emerged as the most prominently explored application vertical, driven by high data sensitivity and institutional silos. \citet{jung2025federatedlearningragintegration} demonstrated that federated RAG pipelines outperform traditional FL setups on clinical question answering tasks by retrieving from a shared medical literature corpus while preserving patient privacy. Complementing this, \citet{jiang2024clinical} introduced a hierarchical retrieval approach that selects relevant hospitals before accessing localized documents, facilitating fine-grained medical inference. \textit{MIRAGE}~\citep{xiong-etal-2024-benchmarking} further formalizes this direction through a clinical QA benchmark tailored for federated evaluation, underscoring the domain's methodological maturity.

Outside of healthcare, \textit{finance and legal} sectors have been identified as strong candidates for Federated RAG due to similar compliance and confidentiality constraints. \textit{C-FedRAG}\citep{addison2024c} targets regulation-aware summarization and document generation, although large-scale evaluation in these domains remains limited. In \textit{enterprise and multilingual QA}, \textit{MKP-QA}\citep{shojaee2025federatedretrievalaugmentedgeneration} introduces a federated benchmark for retrieving internal product knowledge across silos and languages without centralizing sensitive documentation. Meanwhile, \textit{GPT-FedRec}~\citep{zeng2024federatedrecommendationhybridretrieval} explores \textit{personalized recommendation} using a hybrid framework that combines collaborative filtering with retrieval-augmented generation, showing reductions in hallucination and data sparsity.
\begin{table}[t]
\centering\small
\begin{tabular}{l r}
\toprule
\textbf{Application Domain} & \textbf{\# of Studies} \\
\midrule
Healthcare                         & 3 \\
Finance \& Legal                  & 1 \\
Enterprise / Multilingual         & 1 \\
Recommendation / Personal         & 1 \\
General / Multi-domain            & 2 \\
\bottomrule
\end{tabular}
\caption{\small Application domains addressed in federated RAG research. Studies are counted by explicit domain targeting, with several works remaining domain-agnostic or generalized.}
\label{tab:domain}
\vspace{-1.9em}
\end{table}

In addition to these domain-specific efforts, several studies adopt \textit{general or multi-domain} settings. \textit{FeB4RAG}\citep{wang2024} provides a comprehensive evaluation across 16 BEIR-derived tasks, while \textit{FRAG}\citep{zhao2024} focuses on architectural generalization under cryptographic constraints. These studies emphasize benchmarking and scalable deployment over specific industry contexts.

As summarized in Table~\ref{tab:domain}, healthcare dominates the literature with three focused studies, followed by singular works in finance/legal, enterprise/multilingual QA, and recommendation systems. This distribution aligns with real-world concerns over privacy, compliance, and heterogeneous data silos. Earlier Figure~\ref{fig:federatedrag-architecture-taxonomy} illustrated how such deployment settings are reflected in the system-level architecture choices across the literature. While many models are still domain-agnostic, healthcare and enterprise NLP tasks are emerging as leading targets for practical Federated RAG adoption, with future work likely to expand into education, public sector governance, and legal tech.

Beyond these domain categorizations, several privacy-critical scenarios further illustrate the practical relevance of Federated RAG. In healthcare, systems such as C-FedRAG and MIRAGE enable \textit{hospital QA across silos}, allowing institutions to answer clinical questions from shared medical knowledge while ensuring that patient records never leave local storage. In financial services, approaches like RAGRoute support \textit{cross-division portfolio analysis}, where sensitive compliance and investment documents remain within organizational boundaries yet can be queried collaboratively. Finally, in personal and consumer-facing settings, \textit{on-device assistants} equipped with Federated RAG can retrieve from local files or personal notes without exposing them to cloud servers, offering a concrete pathway to privacy-preserving personal AI. 

These scenarios underscore that Federated RAG extends beyond abstract architectural design to address urgent real-world concerns, where the dual imperatives of factual accuracy and data sovereignty are non-negotiable. By grounding retrieval and generation within privacy-preserving frameworks, such applications demonstrate the transformative potential of Federated RAG in bridging regulatory, ethical, and operational requirements.

\subsection{Unified Objective Function and Architecture}
\begin{figure}[t]
    \centering
    \includegraphics[width=0.9\linewidth]{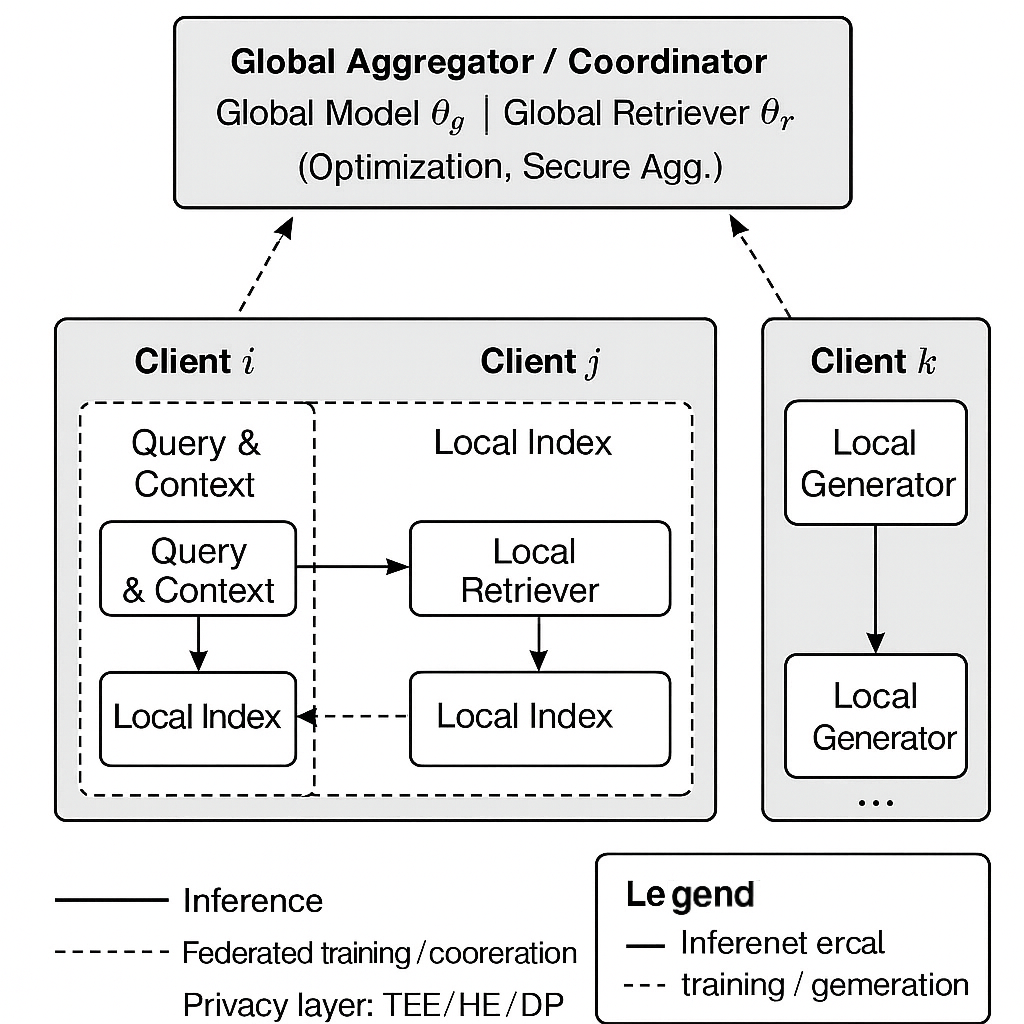}
    \caption{Conceptual architecture of Federated RAG, showing local query processing and index management at each client, coordinated with global model updates.}
    \label{fig:fedrag-architecture}
\end{figure}

While the taxonomy in Figure~\ref{fig:federatedrag-architecture-taxonomy} organizes existing designs across retrieval, generation, and privacy axes, it is also useful to formalize how these components interact within a single optimization framework. We introduce a conceptual objective for Federated RAG see Figure \ref{fig:fedrag-architecture} that captures joint training across clients:

\[
L_{\text{FedRAG}}(\theta) = 
\sum_{i=1}^{M} \tfrac{n_i}{N} 
\big( L_{\text{retrieve}}^{(i)}(\theta_r) + 
      L_{\text{generate}}^{(i)}(\theta_r, \theta_g) \big)
\]

where $M$ is the number of clients, $n_i$ is the number of samples on client $i$, $N$ is the total number of samples, $\theta_r$ are retriever parameters, and $\theta_g$ are generator parameters. This loss highlights that retrieval and generation can be jointly optimized in a federated setting, balancing local personalization with global consistency.

This abstraction complements the taxonomy: retrieval modules may use selective routing or encrypted $k$NN, generation may be centralized, client-specific, or hybrid, and privacy layers such as secure enclaves or differential privacy wrap the pipeline. Together, the unified loss and architecture diagram illustrate how system design choices can be framed within a coherent optimization perspective.

\section{Research Activity and Publication Trends}
Research in Federated Retrieval-Augmented Generation (RAG) has accelerated notably in recent years, driven by rising demand for privacy-preserving NLP and the maturation of retrieval-augmented systems. Before 2022, federated learning (FL) and RAG evolved independently, FL focused on decentralized optimization and privacy, while RAG addressed hallucinations in large language models (LLMs)~\citep{lewis2020rag}. Their integration began in 2022--2023 with early architectural proposals such as \textit{UniMS-RAG}~\citep{chen2022} and \textit{FedE4RAG}~\citep{shin2023}, which outlined how to combine distributed training with retrieval-based augmentation.

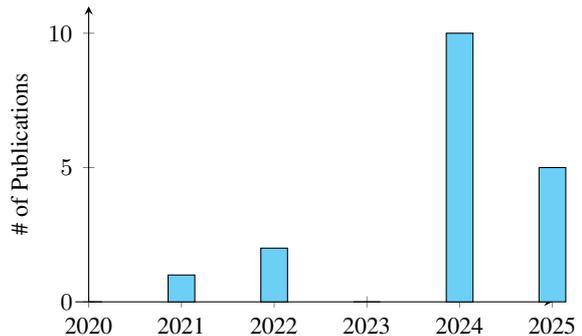
\begin{figure}[t]
\centering
\begin{tikzpicture}
\begin{axis}[
    width=0.48\textwidth,
    height=5.5cm,
    ybar,
    bar width=10pt,
    ymin=0,
    ymax=11,
    axis x line=bottom,
    axis y line=left,
    ylabel={\# of Publications},
    ylabel style={font=\small},
    tick label style={font=\small},
    xtick=data,
    xticklabels={2020, 2021, 2022, 2023, 2024, 2025},
    xticklabel style={rotate=0},
    enlarge x limits=0.15,  
    axis lines=left,
    grid=none,
    major grid style={draw=none},
    minor tick num=0,
    clip=false 
]
\addplot+[ybar, fill=cyan!50, draw=black] coordinates {(0,0) (1,1) (2,2) (3,0) (4,10) (5,5)};
\end{axis}
\end{tikzpicture}
\caption{Publication trends on federated RAG from 2020 to 2025, showing a significant rise in research activity in recent years.}
\label{fig:trends}
\end{figure}

Figure~\ref{fig:trends} shows a sharp rise in 2024, driven by high-stakes LLM use (healthcare/finance) and regulatory pressure. Industry (e.g., NVIDIA’s C-FedRAG~\citep{addison2024c}, Adobe’s MKP-QA~\citep{shojaee2025federatedretrievalaugmentedgeneration}) and open-source toolkits signal consolidation around benchmarks like MIRAGE~\citep{xiong-etal-2024-benchmarking} and FeB4RAG.

Alongside this growth, the field is also beginning to consolidate around shared metrics and benchmarks, which provide clearer evaluation protocols and practical entry points for researchers and practitioners. Notable examples include the \textit{MIRAGE} clinical QA benchmark \citep{xiong-etal-2024-benchmarking}, the \textit{FeB4RAG} federated BEIR benchmark \citep{wang2024}, the \textit{RAGAS} evaluation toolkit for hallucination and consistency \citep{shah2023}, and the open-source \textit{FedRAG toolkit} (Vector Institute). Together, these resources mark an inflection point: Federated RAG research is evolving from isolated system proposals to a field with standardized evaluation pipelines. A comprehensive summary of these toolkits and datasets is provided in Appendix~\ref{tab:fedrag-allstudies}.

\subsection{Privacy \& Security}
Federated RAG must secure both retrieval and generation. \textit{C\mbox{-}FedRAG} confines the entire pipeline to SGX enclaves, shielding queries and documents from untrusted servers. \textit{FRAG} encrypts vector search with IND\mbox{-}CPA--secure homomorphic $k$NN, avoiding raw\mbox{-}index exposure, while \textit{FedE4RAG} performs local augmentation at inference to minimise data movement. These techniques uphold regulations such as ``right\mbox{-}to\mbox{-}be\mbox{-}forgotten'' but add latency and can hamper cross\mbox{-}silo reasoning; future work should formalise threat models and quantify the privacy--utility frontier.

\subsection{Model Adaptation \& Knowledge Management}
Maintaining knowledge without a central store is equally demanding. \textit{FedE4RAG} trains dense retrievers federatively, preserving locality, while adapter\mbox{-}style updates and federated embedding learning cut communication. Hybrid planners like \textit{RAGRoute} select local or remote answers when indices drift. Yet index synchronisation, conflict resolution, and continual adaptation across heterogeneous silos remain open. CRDT\mbox{-}based distributed indexing and meta\mbox{-}learned retriever personalisation are promising, but benchmarks must cover timestamped document shifts, bandwidth ceilings, and real\mbox{-}world constraints.

\section{Applications}
Federated RAG is particularly suited for domains where data sensitivity, personalization, and access to localized knowledge are paramount. In \underline{healthcare}, it enables clinical question answering systems that draw on hospital-specific knowledge bases while ensuring that sensitive patient data remains on-premise, complying with strict privacy regulations. In \underline{financial services}, it facilitates regulation-aware summarization and reporting by incorporating institution-specific documents and compliance guidelines without exposing confidential data externally. Education, enterprise knowledge management, and legal technology also present high-impact opportunities. In \underline{education}, federated RAG can power personalized tutoring systems that align with local curricula and student learning profiles. Enterprise use cases include internal document retrieval and report generation without requiring the centralization of proprietary content. \underline{Legal applications} range from case-specific document analysis to retrieval-augmented reasoning over firm-confidential corpora. 

These examples illustrate the broad utility of federated RAG in supporting \textit{privacy-preserving}, \textit{context-aware}, and \textit{domain-specific} language generation across distributed settings.
\section{Challenges, Gaps and Trends}
In this section, we discuss the overarching challenges, research gaps, and emerging trends for federated RAG. Table~\ref{tab:rq_mapping} summarizes how our research questions (RQ1–RQ4) align with specific challenge areas identified from the 18 primary studies.
\begin{table}[ht]
\small
\centering
\renewcommand{\arraystretch}{1.15}
\begin{tabular}{@{}lp{5.2cm}@{}}
\toprule
\shortstack[l]{\textbf{Research} \\ \textbf{Question}} & \textbf{Key Insights / Challenge Areas} \\
\midrule
\textbf{RQ1} & System heterogeneity in data and clients; need for scalable federated RAG architectures. \\
\textbf{RQ2} & Privacy and security trade-offs; mechanisms for trust and compliance in federated setups. \\
\textbf{RQ3} & Retrieval and generation efficiency; balancing quality against resource costs. \\
\textbf{RQ4} & Evaluation gaps and future directions, such as benchmarks and domain-specific adaptations. \\
\bottomrule
\end{tabular}
\caption{\small Mapping of RQ1–RQ4 to key insights and challenge areas in federated RAG.}
\label{tab:rq_mapping}
\end{table}

\subsection*{Design Patterns and Practical Guidelines}
Beyond high-level challenges, recent studies also surface recurring \textit{design patterns} that can guide implementers in practice:

\begin{itemize}
    \item \textbf{Retrieval.} Systems employ either selective query routing (e.g., RAGRoute) to reduce redundant cross-silo traffic, or encrypted $k$NN search (e.g., FRAG) to ensure confidentiality at the cost of higher latency.
    \item \textbf{Generation.} Approaches range from centralized generation, which improves consistency but raises privacy concerns, to fully client-side generation, which preserves data sovereignty but increases coordination complexity. Hybrid designs (e.g., GPT-FedRec) combine global knowledge with local personalization.
    \item \textbf{Personalization.} Adapter tuning and federated embedding alignment (e.g., FedE4RAG) allow lightweight personalization without overwhelming communication budgets, making them attractive for heterogeneous client settings.
    \item \textbf{Trade-offs.} Every design choice involves balancing privacy, latency, and accuracy. For example, secure enclaves (C-FedRAG) provide strong protection but add overhead; selective routing reduces cost but may risk coverage. These trade-offs highlight the importance of aligning system design with domain-specific regulatory and operational requirements.
\end{itemize}

Taken together, these patterns provide actionable guidance: implementers can view Federated RAG architectures not as ad hoc designs but as configurable combinations of retrieval, generation, and personalization strategies, each with well-understood implications for scalability and compliance.

\noindent\textbf{System Heterogeneity, Scalability, and Performance Trade-offs}~--
Heterogeneous clients introduce latency, uneven resource use, and noisy retrieval results. To address this, recent work explores selective query routing (e.g., \textit{RAGRoute}), benchmark-driven source selection strategies  \textsc{FeB4RAG}, and encrypted caching with probabilistic gating \textit{FRAG}. Together, these methods illustrate how different design choices balance recall, privacy, and latency, highlighting the inherent trade-offs in federated RAG. Overall, such advances directly target the architectural and efficiency concerns that must be solved for scalable deployment (RQ1).

\noindent\textbf{Privacy and Security Considerations}~--
Federated RAG must guarantee data confidentiality throughout both retrieval and generation, making privacy a core concern (RQ2). Traditional RAG pipelines risk information leakage through exposed queries or retrieved documents. Recent systems address this through trusted execution and encryption. \textit{C-FedRAG}, for example, runs indexing and generation entirely inside secure enclaves, protecting sensitive content even when orchestrated by untrusted servers. \textit{FRAG} further strengthens guarantees via homomorphic encryption, enabling encrypted vector search without exposing embeddings or indices—achieving IND-CPA security with acceptable overhead. Compliance with data governance (e.g., right-to-be-forgotten) motivates designs that avoid raw data centralization. \textit{FedE4RAG} exemplifies this by performing local augmentation at inference, ensuring data minimization. However, privacy-preserving techniques often introduce latency and limit cross-silo reasoning, raising the challenge of balancing security with system utility. Future work must formalize threat models and quantify trade-offs between confidentiality, accuracy, and responsiveness in real-world deployments.

\noindent\textbf{Architectural Maturity}~--
Retrieving and updating knowledge without centralized storage is a central challenge for federated RAG systems (RQ3). \textit{FedE4RAG} proposes client-side retriever training via federated learning, improving generalization while respecting data locality. To mitigate resource constraints, current approaches favor parameter-efficient methods such as adapter tuning or federated embedding learning. Hybrid pipelines like \textit{RAGRoute} arbitrate between federated and RAG responses based on confidence, improving robustness when document indices are stale. However, synchronizing indices, resolving data conflicts, and enabling continual adaptation across diverse silos remain open problems. For domain-specific systems, e.g., hospital QA linked to EHR silos, federated RAG must reconcile personalization with evolving knowledge, potentially through distributed indexing, meta-learning, and lightweight model updates. Future work should explore CRDT-based index synchronization and meta-learned retriever personalization to enable continual, privacy-preserving adaptation across heterogeneous silos. Benchmarking these techniques under realistic, timestamped document shifts and resource constraints remains an open evaluation priority.

\noindent\textbf{Evaluation and Benchmarking}~-- Federated RAG is a new topic, hence evaluation frameworks (RQ4) are lacking.  Early studies reused QA criteria or imitated federated environments, making comparisons impossible.  Federated retrieval and creation require standardized datasets and metrics to account for client relevance, privacy constraints, and network costs.  This gap is being filled by recent work:  \textit{FeB4RAG} provides a dataset for federated RAG search, allowing for realistic multi-silo evaluation of retrieval algorithms.  FedRAG may leverage the MIRAGE benchmark (used by RAGRoute) and a new MIMIC-IV-based clinical QA dataset to evaluate systems in certain areas.  Still, evaluating issues exist.  Relevance across clients requires reinterpreting IR measurements like recall or MRR.  Current measurements make it difficult to measure a system's privacy-preserving quality (utility attained for a given level of privacy).  Real-world deployment studies are rare; most are lab-based.  Common evaluation protocols, such as federated versions of popular QA tasks and defined success criteria (accuracy vs. bandwidth vs. privacy), will advance the industry.  Community-driven toolkits (e.g., Flower FedRAG demonstrations) and open-source frameworks are lowering experimental barriers, which are beneficial. Future work should introduce privacy–utility Pareto benchmarks that vary attacker models and bandwidth ceilings, alongside live-stream leaderboards that track adaptation latency and cost as documents drift over time.
\subsection{Outlook and Future Opportunities}Federated RAG research is rapidly evolving, addressing challenges in privacy, scalability, and personalization. Key opportunities include reducing cross-silo latency via neural routing and CRDT-based index synchronization, strengthening privacy through Pareto benchmarks that jointly track attacker models and bandwidth ceilings, and enabling continual adaptation with meta-learned retriever personalization and tiny-ML adapters scheduled to respect heterogeneous devices. Lifelong evaluation resources such as live-stream leaderboards and federated versions of popular QA tasks will clarify trade-offs among accuracy, privacy, and cost as documents drift. Domain-specific optimizations for healthcare, finance, and multi-enterprise settings remain crucial, and community toolkits are lowering experimental barriers. Overall, federated RAG offers a compelling path toward trustworthy, knowledge-grounded AI that operates securely in decentralized environments.
Future research could empirically investigate and quantify these identified architectural trade-offs across diverse real-world federated contexts, thereby enabling more precise architectural recommendations tailored explicitly to domain-specific operational constraints.

\section{Conclusion}
Federated Retrieval-Augmented Generation (RAG) has rapidly moved from concept to practice by pairing privacy-preserving training with knowledge-grounded generation. Our systematic mapping of 18 primary studies charts this evolution, classifying architectures, contributions, and domains, and highlighting a 2024 research surge driven by large-scale language-model adoption and regulatory pressure for confidential data handling.

\textbf{Where it already helps}~-- Early deployments demonstrate clear value in compliance-sensitive settings: \textit{C-FedRAG} answers clinical questions while keeping electronic health-record data in-hospital; \textit{FedE4RAG} powers enterprise knowledge assistants without centralizing proprietary documents; and \textit{RAGRoute} cuts 75\% of redundant cross-silo queries, lowering latency for finance workloads. Across these use cases, federated RAG improves factual grounding, reduces hallucinations, and upholds data-sovereignty requirements that vanilla RAG or centralized LLM serving cannot meet.

\textbf{What has been solved and what has not}~-- Secure enclaves, homomorphic encryption, and differential-privacy mechanisms now offer end-to-end confidentiality; neural routing and probabilistic caching tame bandwidth costs; and federated embedding learning supplies parameter-efficient personalization. Yet four hurdles persist: 
\begin{itemize}
\item \textit{Scalable, CRDT-style index synchronization to keep silos consistent}, 
\item \textit{Meta-learned retriever adaptation that survives concept drift}, 
\item \textit{Privacy–utility evaluation protocols that expose trade-offs under realistic attacker models}, and 
\item \textit{Live benchmarks that measure cost and latency as documents evolve}.
\end{itemize}
\textbf{Why this mapping study matters}~--Federated Retrieval-Augmented Generation has rapidly progressed from early conceptual proposals to a diverse body of system designs, benchmarks, and evaluation resources. Our systematic mapping of 18 primary studies charts this trajectory, identifying common architectural patterns, key challenges, and emerging opportunities across domains such as healthcare, finance, and enterprise knowledge management. 

In moving beyond description, this paper evolves from a static mapping to a \textit{field-shaping foundation}: it combines structured analysis with empirical metrics, introduces theoretical perspectives through a unified objective function, and distills implementation guidance into actionable design patterns. Together, these contributions provide both researchers and practitioners with a clearer roadmap for advancing Federated RAG toward scalable, trustworthy, and domain-ready deployment.

\section*{Limitations}
As with any systematic mapping study, our work is subject to certain limitations that should be acknowledged when interpreting its findings.
\paragraph{External Validity} This study aimed to synthesize trends and insights in Federated RAG research based on a carefully defined corpus of 18 primary studies spanning 2020–2025. To ensure external validity, we explicitly scoped our review to works that integrate retrieval-augmented generation within federated or privacy-preserving systems, across NLP, IR, ML, and security venues. While we employed backward reference tracing and cross-domain search strategies, there remains a possibility that relevant studies, especially those in proprietary industry settings or less commonly indexed venues, may have been missed. Furthermore, the current corpus reflects research trends primarily from Western and East Asian academic and industrial ecosystems, and may not fully represent globally distributed research efforts.
\paragraph{Construct Validity} Federated RAG is an emerging area, and its boundaries are still fluid across communities. Although we adhered to a strict inclusion protocol and used Kitchenham's guidelines for mapping studies, differing interpretations of what constitutes a Federated RAG system (e.g., systems using static knowledge bases vs. dynamic retrieval) may introduce conceptual variance. In some cases, papers labeled as federated learning systems implicitly used retrieval or augmentation mechanisms without explicit reference to RAG paradigms. While we erred on the side of inclusivity, such conceptual ambiguity may influence the completeness or framing of our categories. Moreover, reliance on peer-reviewed and preprint literature means that grey literature, technical documentation, or non-archival systems (e.g., deployed prototypes) were excluded.
\paragraph{Conclusion Validity} Our results stem from a small set of studies, so generalizations should be cautious; the 2024 surge may be a short-term spike. Evolving definitions and tools mean our taxonomy is a snapshot of a rapidly changing landscape.

\paragraph{Other Limitations} 
Despite a comprehensive search, some studies using unconventional terms or industry-only contributions may have been missed, limiting reproducibility. Given the rapid pace of LLM innovation, periodic re-evaluation is needed to keep this mapping relevant.

\bibliographystyle{acl_natbib}

\appendix

\section{Extended Resources and Comparative Synthesis}

\subsection{Benchmarks and Toolkits}
To complement the taxonomy and trends discussed in the main paper, we provide expanded descriptions of benchmarks and toolkits that now support Federated RAG evaluation and deployment:

\begin{itemize}
    \item \textbf{C-FedRAG} \citep{addison2024c}: A confidential QA system executed entirely in Trusted Execution Environments (SGX), demonstrating strong privacy guarantees in clinical domains.
    \item \textbf{RAGRoute} \citep{guerraoui2025efficient}: A query routing framework that reduces redundant cross-silo queries by up to 75\% while maintaining accuracy.
    \item \textbf{FRAG} \citep{zhao2024}: A homomorphic-encryption-based $k$NN retrieval system offering IND-CPA security with latency within 3--5$\times$ of plaintext baselines.
    \item \textbf{GPT-FedRec} \citep{zeng2024federatedrecommendationhybridretrieval}: A hybrid recommendation model combining collaborative filtering with RAG, achieving +5--7\% personalization hit-rate.
    \item \textbf{FedE4RAG} \citep{shin2023}: A modular federated dense retriever framework with homomorphic encryption and distillation for end-to-end privacy.
    \item \textbf{UniMS-RAG} \citep{chen2022}: A unified multi-source RAG system integrating client-side retrieval with a shared generator for adaptive latency control.
    \item \textbf{MIRAGE} \citep{xiong-etal-2024-benchmarking}: A clinical QA benchmark designed for federated hospital data, formalizing evaluation in healthcare contexts.
    \item \textbf{MKP-QA} \citep{shojaee2025federatedretrievalaugmentedgeneration}: A benchmark for enterprise multilingual QA across product silos, supporting federated evaluation in industry.
    \item \textbf{FeB4RAG} \citep{wang2024}: A federated benchmark spanning 16 BEIR-derived domains to study routing strategies and retrieval efficiency.
    \item \textbf{RAGAS} \citep{shah2023}: An evaluation toolkit introducing hallucination, factuality, and robustness metrics tailored to federated RAG pipelines.
    \item \textbf{FedRAG} \citep{addison2024c}: A distributed EHR QA system that introduces hierarchical retrieval across hospitals while preserving patient privacy.
    \item \textbf{RAG} \citep{chen2022}: A medical large language model integrated with RAG, demonstrating gains over FL-only baselines.
    \item \textbf{DP-FedKGE} \citep{peng2021differentially}: A framework for differentially private federated knowledge graph embeddings, enabling reasoning with entity-level privacy.
    \item \textbf{PPDA} \citep{9484437}: A lightweight decentralized aggregation protocol that ensures privacy without relying on DP or HE.
    \item \textbf{DeTrustFL} \citep{xu2022detrust}: A cryptographic trust consensus mechanism preventing disaggregation attacks in federated setups.
    \item \textbf{FedX} \citep{han2022fedx}: An unsupervised federated representation learning system using cross-client knowledge distillation.
   \item \textbf{FL–LLM surveys}~\citep{wu2025survey}: A survey of federated LLM fine-tuning methods, with explicit discussion of RAG integration.
    \item \textbf{FRAD (Zhao)} \citep{zhao2024}: A general-purpose federated architecture demonstrating cryptographic retrieval under multi-domain constraints.
\end{itemize}

Together, these resources provide the beginnings of a shared evaluation layer, enabling more consistent comparisons across systems and domains.

\subsection{Full Comparative Synthesis of Surveyed Studies}
Table~\ref{tab:fedrag-allstudies} extends the synthesis presented in Section~3.1 by aligning each of the 18 Federated RAG studies (2020–2025) with their reported benefits and outcomes. This consolidated view highlights where empirical improvements have been quantified, where cryptographic and privacy guarantees dominate, and where new benchmarks or toolkits have been introduced.


\section{Comparative Synthesis of Federated Studies}
In addition to the qualitative taxonomy presented in Section3.1, we provide a consolidated comparative synthesis of all 18 Federated RAG studies surveyed between 2020–2025 (Table\ref{tab:fedrag-allstudies}). This extended table aligns each system with its reported benefit and empirical or conceptual outcome, thereby offering a more granular view of how diverse contributions map to performance, privacy, efficiency, and evaluation. For instance, system-level advances such as C-FedRAG and RAGRoute quantify measurable improvements in accuracy and communication efficiency, while cryptographic frameworks like FRAG demonstrate the latency trade-offs inherent in strong privacy guarantees. Benchmarking and evaluation initiatives, including FeB4RAG, MIRAGE, and RAGAS, highlight the field’s growing emphasis on standardized protocols, whereas systems like FedE4RAG and UniMS-RAG focus on architectural integration. Early foundational efforts (e.g., DP-FedKGE, PPDA, DeTrustFL) continue to shape privacy-preserving mechanisms, while recent domain-specific benchmarks (e.g., MKP-QA) expand application horizons. Taken together, this synthesis complements the main text by grounding the survey’s classifications in concrete evidence and by illustrating the breadth of technical strategies underpinning Federated RAG research.
\begin{table*}[t]
\centering
\small
\setlength{\tabcolsep}{6pt} 
\renewcommand{\arraystretch}{1.15}
\begin{tabular}{p{7.5cm} p{6.5cm}}
\toprule
\textbf{System + Reported Benefit} & \textbf{Outcome / Contribution} \\
\midrule
\textbf{C-FedRAG} ~\citep{addison2024c} \\ 
\quad Clinical QA in trusted enclaves  
  & +12.7\% QA accuracy (59.8→72.5) using SGX \\
\textbf{RAGRoute} \citep{guerraoui2025efficient} \\ 
\quad Adaptive query routing across silos  
  & –75\% redundant queries with 72\% accuracy \\
\textbf{FRAG} \citep{zhao2024} \\ 
\quad Encrypted kNN retrieval (HE-based)  
  & 3–5$\times$ latency compared to plaintext \\
\textbf{GPT-FedRec} \citep{zeng2024federatedrecommendationhybridretrieval} \\ 
\quad Personalized recommendation  
  & +5–7\% hit rate improvement \\
\textbf{FedE4RAG} \citep{shin2023} \\ 
\quad Federated dense retrievers + HE distillation  
  & End-to-end privacy-preserving retrievers \\
\textbf{UniMS-RAG} \citep{chen2022} \\ 
\quad Multi-source retrieval + shared generator  
  & Unified retrieval-control, adaptive latency \\
\textbf{MIRAGE} \citep{xiong-etal-2024-benchmarking} \\ 
\quad Clinical QA benchmark  
  & First federated benchmark for clinical QA \\
\textbf{MKP-QA} \citep{shojaee2025federatedretrievalaugmentedgeneration} \\ 
\quad Enterprise multilingual QA benchmark  
  & Cross-silo multilingual evaluation \\
\textbf{FeB4RAG} \citep{wang2024} \\ 
\quad Federated BEIR benchmark  
  & Evaluation across 16 BEIR-derived domains \\
\textbf{RAGAS} \citep{shah2023} \\ 
\quad Evaluation toolkit for hallucination/consistency  
  & Metrics for hallucination, evidence quality, robustness \\
\textbf{FedRAG} \citep{addison2024c} \\ 
\quad Distributed EHR QA  
  & Hierarchical hospital retrieval pipeline \\
\textbf{RAG} \citep{chen2022} \\ 
\quad Medical LLM + RAG integration  
  & Outperforms FL-only baselines on clinical QA \\
\textbf{DP-FedKGE} \citep{peng2021differentially} \\ 
\quad Differentially private embeddings  
  & Entity-level DP protection for KG embeddings \\
\textbf{PPDA} \citep{9484437} \\ 
\quad Lightweight decentralized aggregation  
  & Privacy without DP/HE overhead \\
\textbf{DeTrustFL} \citep{xu2022detrust} \\ 
\quad Cryptographic trust consensus  
  & Prevents disaggregation attacks in FL \\
\textbf{FedX} \citep{han2022fedx} \\ 
\quad Cross-client knowledge distillation  
  & Unsupervised federated representation learning \\
\textbf{FL–LLM surveys}~\citep{wu2025survey} \\ 
\quad Survey of FL-LLM fine-tuning  
  & Overview, references to RAG methods \\
\textbf{FRAD (Zhao)} \citep{zhao2024} \\ 
\quad General multi-domain architecture  
  & Cryptographic constraints, scalable retrieval \\
\bottomrule
\end{tabular}
\caption{Comparative synthesis of 18 Federated RAG studies (2020–2025). Each system (bold, first line) is paired with its reported benefit (indented line) and the corresponding outcome or contribution.}
\label{tab:fedrag-allstudies}
\end{table*}

\end{document}